\documentclass[10pt, a4paper, conference, compsocconf, twocolumn]{IEEEtran}
\usepackage{ textcomp }

\usepackage{hyperref}       

\usepackage{url}            

\usepackage{booktabs}       

\usepackage{amsfonts}       

\usepackage{nicefrac}       

\usepackage{microtype}      

\usepackage{graphicx}

\usepackage{subcaption}

\usepackage{amsmath}

\usepackage{algpseudocode}

\usepackage{algorithm}
	
\algnewcommand\algorithmicforeach{\textbf{for each}}

\algdef{S}[FOR]{ForEach}[1]{\algorithmicforeach\ #1\ \algorithmicdo}

\begin{document}
	
	\title{Handwriting Recognition of Historical Documents with few labeled data}
	
	\author{\IEEEauthorblockN{Edgard Chammas and Chafic Mokbel}
		\IEEEauthorblockA{University of Balamand\\
			El-Koura, Lebanon\\
			\{edgard,chafic.mokbel\}@balamand.edu.lb}
		\and
		\IEEEauthorblockN{Laurence Likforman-Sulem}
		\IEEEauthorblockA{Institut Mines Telecom, Telecom ParisTech and Universit\'e Paris-Saclay\\
			Paris, France\\
			laurence.likforman@telecom-paristech.fr}
	}

	\maketitle
	
	\begin{abstract}
		Historical documents present many challenges for offline handwriting recognition systems, among them, the segmentation and labeling steps. Carefully annotated text-lines are needed to train an HTR system. In some scenarios, transcripts are only available at the paragraph level with no text-line information. In this work, we demonstrate how to train an HTR system with few labeled data. Specifically, we train a deep convolutional recurrent neural network (CRNN) system on only 10\% of manually labeled text-line data from a dataset and propose an incremental training procedure that covers the rest of the data. Performance is further increased by augmenting the training set with specially crafted multi-scale data. We also propose a model-based normalization scheme which considers the variability in the writing scale at the recognition phase. We apply this approach to the publicly available READ dataset{\protect\NoHyper\footnote{https://read.transkribus.eu/}\protect\endNoHyper}. Our system achieved the second best result during the ICDAR2017 competition {\protect\NoHyper\cite{sanchez2017icdar2017}\protect\endNoHyper}.
	\end{abstract}
	
	
	\IEEEpeerreviewmaketitle
	
	\section{Introduction}
	\label{section1}
\begin{figure}[h]
		\centering
		\begin{subfigure}[b]{0.24\textwidth}
			\centering
			\includegraphics[width=\linewidth]{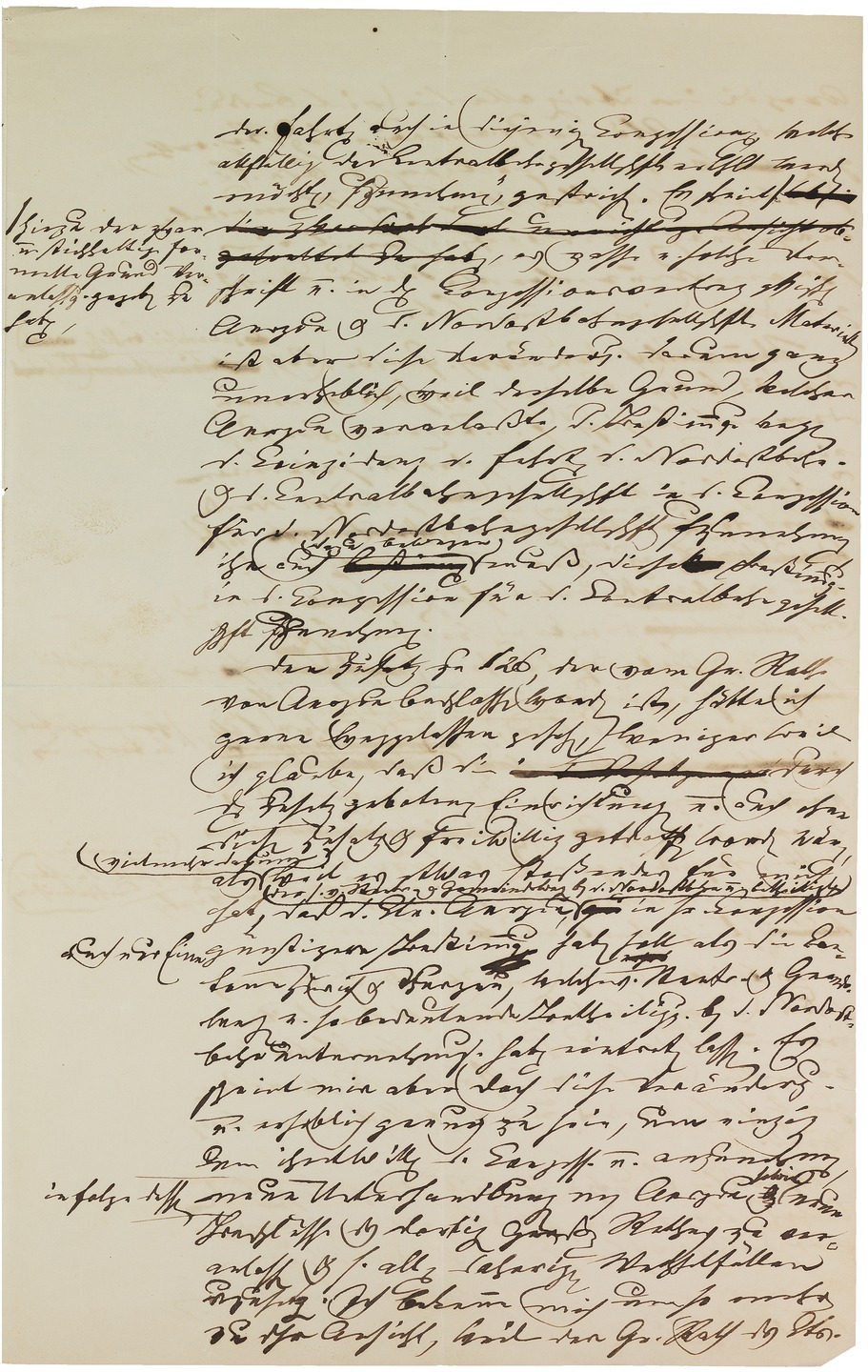}
		\end{subfigure}\hfill
		\begin{subfigure}[b]{0.24\textwidth}
			\centering
			\includegraphics[width=\linewidth]{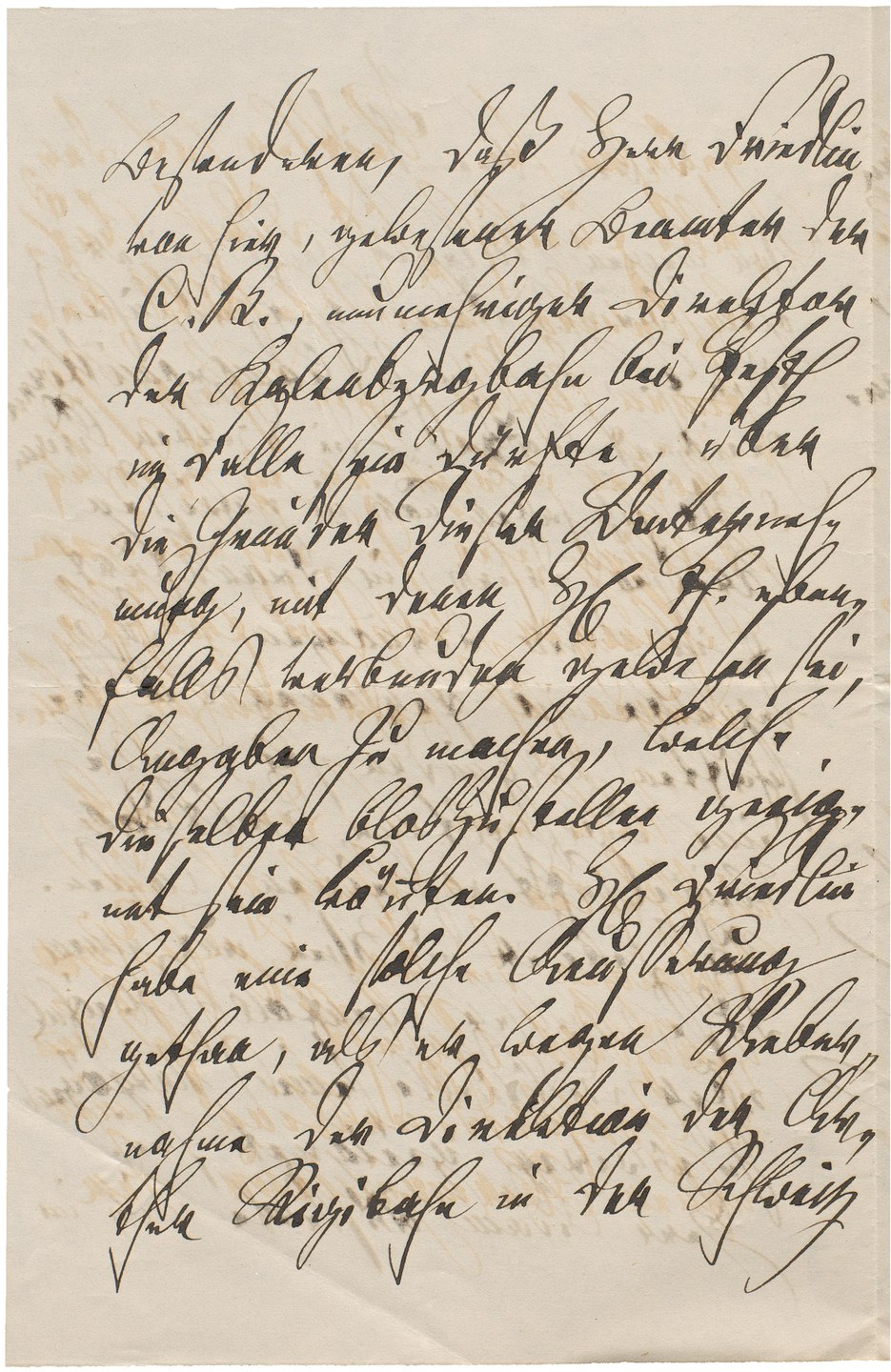}
		\end{subfigure}\hfill
		\caption{Old manuscripts from the READ 2017 dataset.}\label{fig:figure1}
	\end{figure}

	Most state-of-the-art offline handwriting text recognition (HTR) systems work at the line level by transforming the text-line image into a sequence of feature vectors. These features are fed into an optical model (e.g, recurrent neural network) in order to recognize the handwritten text. Recent work on text detection and localization {\protect\NoHyper\cite{moysset2017full}\protect\endNoHyper} at the document level, and joint line segmentation and recognition at the paragraph level {\protect\NoHyper\cite{bluche2016joint}\protect\endNoHyper} showed promising results. However, the best recognition results are still achieved by the systems working at the line level {\protect\NoHyper\cite{voigtlaender2016handwriting}\protect\endNoHyper}. The automatic segmentation of paragraphs into lines is even more challenging on historical documents. Old manuscripts are often acquired as low resolution images with degraded quality, with overlapping characters across adjacent text-lines (see figure {\protect\NoHyper\ref{fig:figure1}\protect\endNoHyper}). Supervised (or at least semi-supervised) paragraph segmentation is needed to label each text-line in order to train an HTR system. However, this is a tedious and time consuming task that is not always feasible for different reasons (budget, time, priority, and availability of text data). When transcriptions are primarily provided at the paragraph level, the first challenge consists in aligning the training transcription data with the corresponding lines in the image. In this paper, we propose to perform such an alignment after training a first recognition system on a limited amount of annotated data. The first system serves to bootstrap the whole process. We also suggest to augment the amount of data by generating multiscale synthetic data in order to better consider the scale factor in the test images. We apply this approach to the READ dataset, a multilingual Latin offline handwriting dataset. The training data provided during the ICDAR2017 competition{\protect\NoHyper\footnote{https://scriptnet.iit.demokritos.gr/competitions/8/}\protect\endNoHyper} were part of the Alfred Escher Letter Collection (AEC), with a large vocabulary of more than 130k words. The test data were letter documents from the same period of AEC. In section {\protect\NoHyper\ref{section2}\protect\endNoHyper}, we present our state of the art deep convolutional recurrent neural network (CRNN) that we used in ICDAR2017 competition on handwritten text recognition. During the competition, 10000 pages were available for training with transcriptions provided at the paragraph level only. In section {\protect\NoHyper\ref{section3}\protect\endNoHyper}, we demonstrate how to train an HTR system by using a small amount of manually segmented and labeled text-lines to create a bootstrap model. We further improve the performance of our system by augmenting the training set with specially crafted synthetic data, explicitly taking into consideration the variability in the writing scale (section {\protect\NoHyper\ref{section4}\protect\endNoHyper}). In section {\protect\NoHyper\ref{section5}\protect\endNoHyper}, we propose a model-based normalization scheme that considers the writing scale variability in the test data. Our system achieved the second best result during the ICDAR2017 competition.
	
	\section{CRNN system description}
	
	\label{section2}
	Our system is a deep Convolutional Recurrent Neural Network (CRNN) inspired from the VGG16 architecture {\protect\NoHyper\cite{simonyan2014very}\protect\endNoHyper} used for image recognition. We use a stack 13 convolutional ($3\times3$ filters, $1\times1$ stride) layers followed by three Bidirectional LSTM layers with 256 units per layer. Each LSTM unit has one cell with enabled peephole connections. Spacial pooling (max) is employed after some convolutional layers. To introduce non-linearity, the Rectified Linear Unit (ReLU) activation function was used after each convolution. It has the advantage of being resistant to the vanishing gradient problem while being simple in terms of computation, and was shown to work better than sigmoid and tanh activation functions {\protect\NoHyper\cite{gu2015recent}\protect\endNoHyper}. A square shaped sliding window is used to scan the text-line image in the direction of the writing. The height of the window is equal to the height of the text-line image, which has been normalized to 64 pixels. The window overlap is equal to 2 pixels to allow continuous transition of the convolution filters. For each analysis window of $64\times64$ pixels in size, 16 feature vectors are extracted from the feature maps produced by the last convolutional layer and fed into the observation sequence. It is worth noting that the amount of feature vectors extracted from each sliding windows is important. The number must be reasonable as to provide a good sampling for the image. Based on previous experiments, we found out that oversampling (32 feature vectors per window) and under-sampling (8 feature vectors per window) will decrease the performance. Sixteen feature vectors were found to work best for our architecture. Since for each of the 16 columns of the last 512 feature maps, the columns of height 2 pixels are concatenated into a feature vector of size 1024 ($512\times2$).\newline \newline Thanks to the CTC objective function {\protect\NoHyper\cite{graves2006connectionist}\protect\endNoHyper}, the system is end-to-end trainable. The convolutional filters and the LSTM units weights are thus jointly learned within the back-propagation procedure. We chose to keep the network simple with a relatively small number of parameters. We thus combine the forward and backward outputs at the end of the BLSTM stack {\protect\NoHyper\cite{zeyer2016towards}\protect\endNoHyper} rather than at each BLSTM layer. We also chose not to add additional fully-connected layers. The LSTM unit weights were initialized as per {\protect\NoHyper\cite{glorot2010understanding}\protect\endNoHyper} method, which proved to work well and helps the network convergence faster. This allows the network to maintain a constant variance across the network layers which keeps the signal from exploding to a high value or vanishing to zero. \newline \newline The weight matrix $ W_{ij} $  were initialized with a uniform distribution given as $ W_{ij} \sim U(- \frac{{\sqrt{6}}}{n}, \frac{{\sqrt{6}}}{n} ) $, where $ n $ is the total number of input and output neurons at the layer (assuming all layers are of the same size). \newline \newline Adam optimizer {\protect\NoHyper\cite{kingma2014adam}\protect\endNoHyper} was used to train the network with initial learning rate of 0.001. This algorithm could be thought of as an upgrade for RMSProp {\protect\NoHyper\cite{tieleman2012lecture}\protect\endNoHyper}, offering bias correction and momentum {\protect\NoHyper\cite{qian1999momentum}\protect\endNoHyper}. It provides adaptive learning rates for the stochastic gradient descent update computed from the first and second moments of the gradients. It also stores an exponentially decaying average of the past squared gradients (similar to Adadelta {\protect\NoHyper\cite{zeiler2012adadelta}\protect\endNoHyper} and RMSprop) and the past gradients (similar to momentum). Batch normalization as described in {\protect\NoHyper\cite{ioffe2015batch}\protect\endNoHyper}, was added after each convolutional layer in order to accelerate the training process. It basically works by normalizing each batch by both mean and variance. The network was trained in an end-to-end fashion with the CTC loss function {\protect\NoHyper\cite{graves2006connectionist}\protect\endNoHyper}. A token passing algorithm was used for decoding {\protect\NoHyper\cite{fischer2012handwriting}\protect\endNoHyper}. It integrates a bigram language model with modified Kneser-Ney discounting {\protect\NoHyper\cite{chen1996empirical}\protect\endNoHyper}, built from the available training data. It is worth noting that no preprocessing is needed. The system works directly on raw images. The full architecture is provided at the end of this paper (figure {\protect\NoHyper\ref{figure5}\protect\endNoHyper}) and the code can be found on GitHub{\protect\NoHyper\footnote{https://github.com/0x454447415244/HandwritingRecognitionSystem}\protect\endNoHyper}.

	\section{Incremental training with few labeled data}
	
	\label{section3}
	
\begin{figure}[h]
		
		\centering
		
		\includegraphics[width=1.0\linewidth]{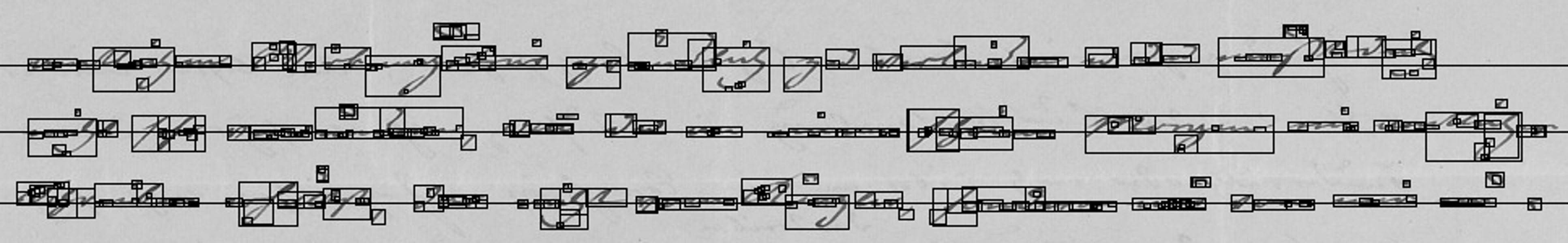}
		
		\caption{Candidate baselines with contours bounding boxes.}
		
		\label{figure1}
		
	\end{figure}

	With no line information provided, few labeled text-lines are needed to bootstrap the training process. We used an automatic segmentation algorithm to extract line images from the document images. The algorithm selects candidate baselines by analyzing contours distribution. It then assigns each contour to one of the baselines based on a number of criteria, related to the average distance between two lines and the distance between the contour center and the line (see figure {\protect\NoHyper\ref{figure1}\protect\endNoHyper}). Only 10\% of the pages were manually verified, making sure the line segmentation is correct, and used to bootstrap the training process. Besides the 10,000 training pages, 50 annotated pages at the line level were provided during the competition and were used for validation in the training process. The initial recognition system, trained on 10\% of the data, achieved 9.2\% raw label error rate (LER). This performance can be considered good enough to allow an incremental training of the network from the rest of the data.

	As a next step, the system was set to recognize the remaining 90\% of the segmented line images in the training set. The recognized lines were mapped to lines in the ground-truth data for each page, based on the Levenshtein distance {\protect\NoHyper\cite{levenshtein1966binary}\protect\endNoHyper} between the text lines. A mapping is considered valid when the edit distance is less than or equal to half the length of the reference line. Following this process, and according to this threshold on the Levenshtein distance, 80\% of the available text-lines were selected to retrain the system, while the rest (20\%) were discarded. The retrained system achieved a relative decrease of 20\% in raw LER on the validation set (see table {\protect\NoHyper\ref{table1}\protect\endNoHyper}). The process could have been restarted after having trained the system with the new data, or even iterated. An improved recognition performance could have recovered more training lines. However, we have noticed that most of the discarded line images in the first iteration resulted from wrong segmentation (e.g., two text-lines in a single image, cropped text-line, etc), due to the fact that the algorithm is sensitive to the writing skew. Therefore, more advanced segmentation algorithms are needed to improve the selection/training process, like the ones based on Seam Carving technique {\protect\NoHyper\cite{arvanitopoulos2014seam}\protect\endNoHyper} and dynamic programming, which would have resulted in fewer segmentation errors and therefore more labeled training data. The whole process can be summarized at the end of this paper (algorithm {\protect\NoHyper\ref{algorithm1}\protect\endNoHyper}).
	
		\begin{table}[h]
		
		\caption{System performance on the validation set with different amount of training data.}
		
		\label{table1}
		
		\centering
		
		\begin{tabular}{ccc}
			
			\toprule
			
			System        & Number of text-lines & Label Error Rate (LER) \\
			
			\midrule
			
			10\% training data 	  	 & \texttildelow20k	 & 9.2\% \\
			
			80\% training data   	 & \texttildelow160k & 7.4\%	\\

			\bottomrule
			
		\end{tabular}
		
	\end{table}

	\section{Integration of multi-scale training data}
	
	\label{section4}

	To further enhance the performance of the system, we exploited the variability in the writing scale to augment the training set with text-line images at multiple scales. Based on a vertical scale score {\protect\NoHyper\cite{chammas2016exploitation}\protect\endNoHyper},  the training lines were first classified into 3 classes (Large, Medium and Small) via Jenks natural breaks optimization algorithm {\protect\NoHyper\cite{jenks1967data}\protect\endNoHyper}. By dividing the training set over the three classes, the data volume per class become smaller. To address this problem, we expanded the training set for each class by adding synthetic data resulting from scaling the other classes’ data. For example, we reduce the large images and stretch the small ones (by a predetermined factor for each class) to expand the number of medium sized images. Or we reduce the medium and large sized images to extend the set of small images, etc. To calculate the scaling factors by which a certain image of a given scale class is enlarged or reduced, the average scale measurement score is calculated on the data. For instance, to transform an image $I$ of class $X$ to an image $J$ of class $Y$, we scale $I$ by $E(Y)/E(X)$, where E(X) and E(Y) are the average scale score values for class $X$ and $Y$ respectively. We retrained the baseline system on multi-scale data for one epoch and achieved a 6.5\% raw LER; a relative improvement by 12\% from the previous system.
	
	\begin{figure}[h]
		
		\centering
		
		\includegraphics[width=1.0\linewidth]{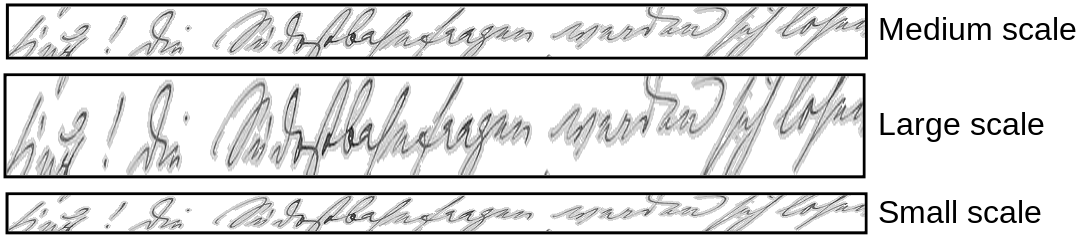}
		
		\caption{An example from the READ dataset where a text-line classified as Medium scale is transformed into a Large and Small scale versions.}
		
		\label{figure3}
		
	\end{figure}
	
	\section{Model-based normalization scheme}
	
	\label{section5}

	To further improve the performance, we proposed to consider the variability in the writing scale in a model-based normalization scheme, where the test data are equalized in order to best fit the core model. In general, consider the recognition phase where a test image characterized by a specific variability is provided at the input of a system trained on a general training set. According to the statistical decision theory, the recognition task identifies the most likely word sequence given the observations as:
	
	\begin{equation}\hat{s} = \underset{s}{\operatorname{arg\,max}} Pr(s \vert \underline{\underline{X}})\end{equation}
	
	where $s$ represents a word sequence, and $ \underline{\underline{X}} $ the observation sequence. To cope with a variability factor $ \theta $ in a test image, it is supposed that a transformation $ T_\theta(.) $ exists with contextual parameter vector $ \theta $ permitting to reduce this variability to a minimum. It is assumed that this parameter is hidden and cannot be measured. A normalized version of the input image $ \underline{\underline{X}} $ can be defined as:
	
	\begin{equation}\widehat{\underline{\underline{X}}} = T_\theta(\underline{\underline{X}})\end{equation}
	
\begin{figure}[h]
		
		\centering
		
		\includegraphics[width=1.0\linewidth]{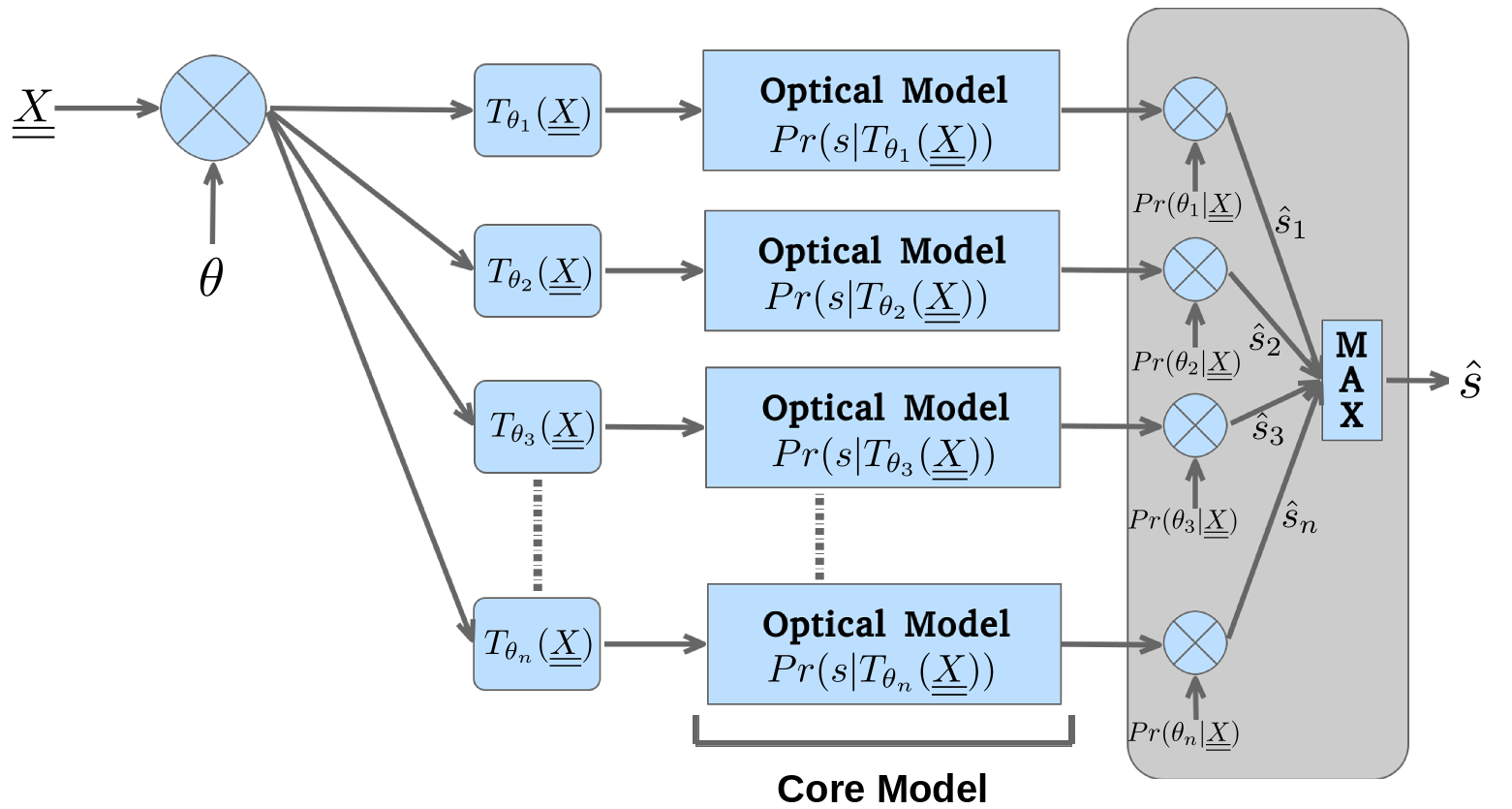}
		
		\caption{Model-based normalization scheme.}
		
		\label{figure4}
		
	\end{figure}

\begin{figure}[h]
	
	\centering
	
	\includegraphics[width=1.0\linewidth]{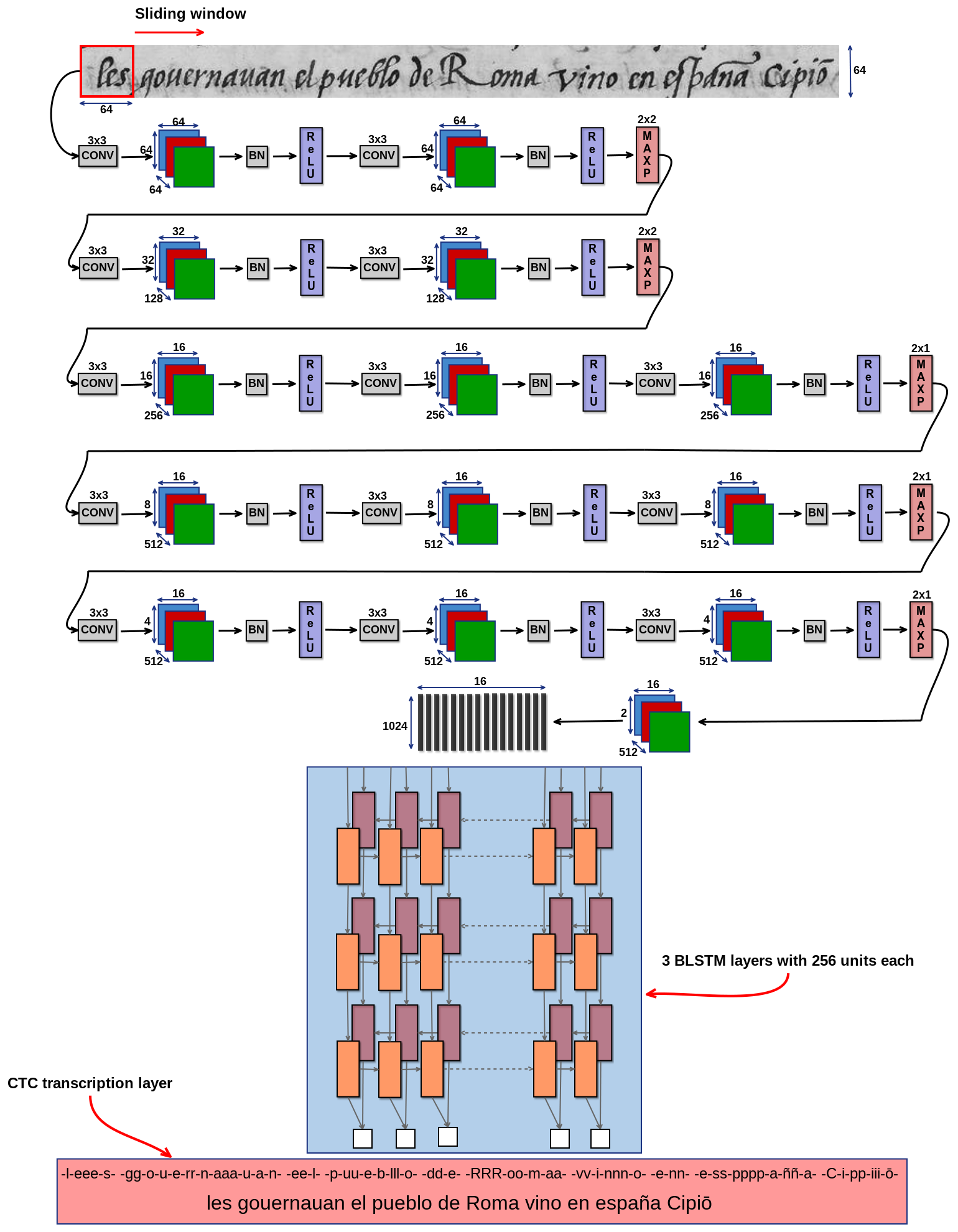}
	
	\caption{Recognition system.}
	
	\label{figure5}
	
\end{figure}

	Assuming the contextual parameter vector $ \theta $ belongs to a finite set, equation 1 can integrate the normalization defined in equation 2 to become:\begin{equation}
	\begin{split}
	\hat{s} & = \underset{s}{\mathrm{argmax}}\sum_{\theta}^{} Pr(s, \theta \vert \underline{\underline{X}}) \\
	& = \underset{s}{\mathrm{argmax}}\sum_{\theta}^{} Pr(s \vert \theta, \underline{\underline{X}}) Pr(\theta \vert \underline{\underline{X}}) \\
	& = \underset{s}{\mathrm{argmax}}\sum_{\theta}^{} Pr(s \vert T_{\theta}(\underline{\underline{X}})) Pr(\theta \vert \underline{\underline{X}})
	\end{split}
	\end{equation}
	For all possible normalizations of the input $ \underline{\underline{X}} $, the system produces solutions with the corresponding scores, considered as posterior probabilities. A combination of the scores permits to re-select the optimal solution (see figure {\protect\NoHyper\ref{figure4}\protect\endNoHyper}). This is considered as an approximation of the right-hand term of equation 3.
	
	We generated multiple versions of the test data by vertically scaling each text-line image to multiple scales (0.7, 0.8,..., 1.3). By considering equation 3, we could write:
	
	\begin{equation} \hat{s} = \underset{s}{\mathrm{argmax}}\sum_{\theta=0.7}^{1.3} Pr(s \vert T_{\theta}(\underline{\underline{X}})) Pr(\theta \vert \underline{\underline{X}}) \end{equation}
	
	We approximate equation 4 by the means of ROVER method {\protect\NoHyper\cite{fiscus1997post}\protect\endNoHyper}. The combination of the recognition scores of the different normalized versions of the test image has yielded to a relative improvement of 14\% in WER from the baseline system. In Table {\protect\NoHyper\ref{table2}\protect\endNoHyper}, we provide the word error rate (WER) and character error rate (CER) obtained with the different systems along with the result of the BYU (Computer Science Department) team who won the first place during the competition. The results show the significant increase in performance using the incremental training of our CRNN system. They also show a significant improvement when better considering the variability of writing scale. Finally, our best system achieves comparable results with the system ranked first in the contest. With 5.5\% running OOV words {\protect\NoHyper\cite{sanchez2017icdar2017}\protect\endNoHyper}, we believe the main difference in performance can be explained by our use of a bigram word language model. It is worth noting that our results can further be improved by using a more performant segmentation, which would also leads to more training data.
	
	\begin{table}[h]
		
		\caption{Effect of multi-scale data on the performance.}
		
		\label{table2}
		
		\centering
		
		\begin{tabular}{ccc}
			
			\toprule
			
			System & CER & WER     \\
			
			\midrule
			
			CRNN (1) 	& 9.18\%  & 25.07\%			\\
			
			CRNN retrained with multi-scale data (2)  & 7.95\%  & 23.09\%			\\
			
			(2) + model-based normalization scheme & 7.74\%  & 21.58\%			\\
			
			\textit{BYU System} & 7.01\%  & 19.06\%			\\

			\bottomrule
			
		\end{tabular}
		
	\end{table}
	
	\section{Conclusions and perspectives}
	
	\label{section6}
	
	In this work, we presented a state-of-the-art CRNN system for text-line recognition of historical documents. We showed how to train such system with few labeled text-line data. Specifically, we proposed to bootstrap an incremental training procedure with only 10\% of manually labeled text-line data from the READ 2017 dataset. We also improved the performance of the system by augmenting the training set with specially crafted synthetic data at multi-scale. At the end, we proposed a model-based normalization scheme by introducing the notion of the variability in the writing scale to the test data. The combination of the multi-scale trained system results on multi-scale test data has yielded the best result. Our system achieved the second position in ICDAR2017 competition, with comparable performance to the winning system, while noting that the overall performance depends on both segmentation and recognition tasks. Our results can be improved by improving the segmentation algorithm which will permit to use more training data. Despite the complex network architecture, we noticed the large impact of the variability in the writing scale on the performance. As a future work, we will be looking into the possibilities for integrating this variability in the modeling. Possibly via an attention mechanism.
	
	\section{Acknowledgment}
	
	\label{section7}
	
	We gratefully acknowledge the support of NVIDIA Corporation with the donation of the Titan Xp GPU used for this research.
	
	\bibliographystyle{IEEEtran}
	\bibliography{ref}
	
	\bigskip
	
    \begingroup
		\let\clearpage\relax 
		\onecolumn 
	\endgroup

\begin{algorithm}
	
	\caption{Incremental alignment process}
	
	\label{algorithm1}
	
	\begin{algorithmic}
		
		\Require $ TrainSet $: Set of all training pages
		
		\Require $ RefText $: Ground-truth text paragraph for each page
		
		\ForEach { page $ P $ in $ TrainSet $ }
		
		\State $ Lines[] \gets Segment(P) $
		
		\State $ RefLineIndex \gets 0 $
		
		\ForEach { line $ L $ in $ Lines $ }
		
		\State $ RecSeq \gets Recognize(L) $
		
		\While {$ RefLineIndex < length(RefText[P]) $}
		
		\State $ RefSeq \gets RefText[P][RefLineIndex] $
		
		\State $ EditDistance \gets Levenshtein(RecSeq, RefSeq) $
		
		\If {$ EditDistance < 0.5 \times length(RefSeq) $}
		
		\State $ Map(L, RefSeq) $
		
		\State $ RefLineIndex \gets RefLineIndex + 1 $
		
		\EndIf
		
		\EndWhile
		
		\EndFor
		
		\EndFor
		
	\end{algorithmic}
	
\end{algorithm}

	\onecolumn

\end{document}